\documentclass[letterpaper]{article} 
\usepackage{aaai2026}  
\usepackage{times}  
\usepackage{helvet}  
\usepackage{courier}  
\usepackage[hyphens]{url}  
\usepackage{graphicx} 
\usepackage{natbib}  
\usepackage{caption} 
\usepackage{algorithm}
\usepackage{algorithmic}
\usepackage{amsfonts}
\usepackage{multirow}
\usepackage{booktabs}

\usepackage{newfloat}
\usepackage{listings}
\DeclareCaptionStyle{ruled}{labelfont=normalfont,labelsep=colon,strut=off} 
\floatstyle{ruled}
\newfloat{listing}{tb}{lst}{}
\floatname{listing}{Listing}
\nocopyright

\title{LidarPainter: One-Step Away From Any Lidar View To Novel Guidance}
\author{
    Yuzhou Ji\textsuperscript{\rm 1,2},
    Ke Ma\textsuperscript{\rm 1},
    Hong Cai\textsuperscript{\rm 2}\thanks{Project Leader},
    Anchun Zhang\textsuperscript{\rm 2},
    Lizhuang Ma\textsuperscript{\rm 1},
    Xin Tan\textsuperscript{\rm 3}\thanks{Corresponding Author}
}
\affiliations{
    \textsuperscript{\rm 1}Shanghai Jiao Tong University, Shanghai, China\\
    \textsuperscript{\rm 2}51WORLD, Shanghai, China\\
    caihong@51sim.com\\
    \textsuperscript{\rm 3}East China Normal University, Shanghai, China\\
    xtan@cs.ecnu.edu.cn
}

\usepackage{bibentry}

\begin{document}

\maketitle

\begin{abstract}
Dynamic driving scene reconstruction is of great importance in fields like digital twin system and autonomous driving simulation.
However, unacceptable degradation occurs when the view deviates from the input trajectory, leading to corrupted background and vehicle models.
To improve reconstruction quality on novel trajectory, existing methods are subject to various limitations including inconsistency, deformation, and time consumption.
This paper proposes LidarPainter, a one-step diffusion model that recovers consistent driving views from sparse LiDAR condition and artifact-corrupted renderings in real-time, enabling high-fidelity lane shifts in driving scene reconstruction.
Extensive experiments show that LidarPainter outperforms state-of-the-art methods in \textbf{speed}, \textbf{quality} and \textbf{resource efficiency}, specifically 7 $\times$ faster than StreetCrafter with only one fifth of GPU memory required.
LidarPainter also supports stylized generation using text prompts such as ``foggy'' and ``night'', allowing for a diverse expansion of the existing asset library.
\end{abstract}
\begin{links}
\link{Proj Page}{https://george-attano.github.io/LidarPainter}
\end{links}


\section{Introduction}

In the fields of digital twin systems and autonomous driving simulation, the reconstruction of dynamic driving scenes is of great significance.
While it is impossible to capture every single driving scenario in reality, a reconstructed scene can not only be edited to create unlimited test cases for evaluating driving algorithms at low costs, but can also unify the data collected on different sensors to build a data-consistent asset library.
However, real-world driving data cannot provide the same surround view input as in static indoor scenes, causing significant degradation as the viewpoint deviates from the original trajectory.
Therefore, obtaining usable reconstructed assets with comprehensive supervision for driving simulation remains a vital challenge.

\begin{figure}[t]
   \centering
   \includegraphics[width = 0.9\columnwidth]{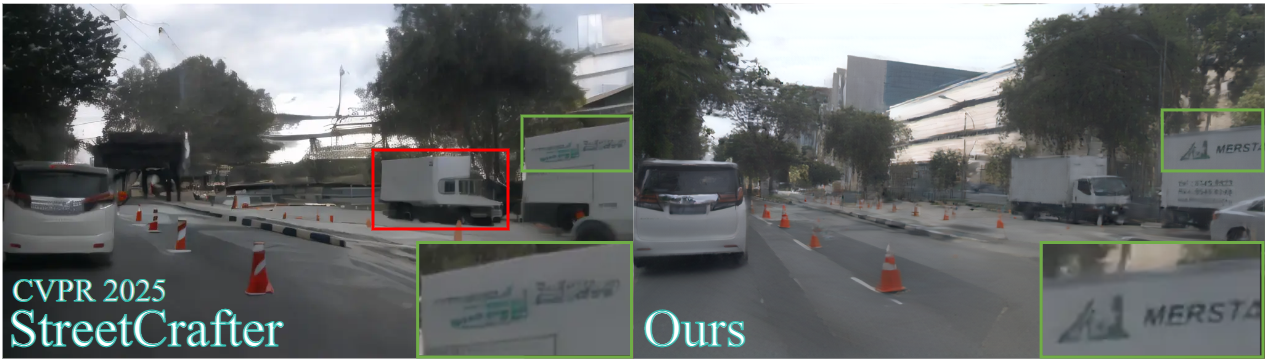}
   \caption{Comparison of diffusion guidance sampled on 7000 iterations. Our method shows much better fidelity and consistency with clear characters on truck, while StreetCrafter generates corrupted vehicles and text.}
   \label{fig:intro}
\end{figure}

To improve reconstruction quality, previous methods \cite{yang2024driving, ni2025recondreamer, zhao2025recondreamer++, yu2025sgd, zhao2025drivedreamer4d} prove the necessity of leveraging generative priors \cite{gao2024vista, wang2024drivedreamer, zhao2025drivedreamer} as guidance.
Similarly, the recent state-of-the-art method, StreetCrafter \cite{yan2025streetcrafter}, trained a video diffusion model based on Vista \cite{gao2024vista}, successfully generating novel view sequences with precise camera control under LiDAR conditions.
However, StreetCrafter degrades largely on sparse LiDAR renderings, corrupting the geometric structure of objects and failing in most details, such as text and patterns (as shown in Figure \ref{fig:intro}), which limits the realism of reconstructed assets. 
StreetCrafter takes up a good deal of GPU memory, specifically 55 GB for a clip of only 15 frames.
This also means a long driving sequence needs to be processed in fragments, resulting in sudden changes of scene architecture on the junction of video clips and causes artifacts.
Moreover, a three-stage video diffusion sample in StreetCrafter of only 100 frames can take up to 45 minutes on a single A100 GPU, even more than the complete training time for scene reconstruction. 

By analyzing existing methods, we believe that there is a common misunderstanding in improving driving scene reconstruction quality.
While the consistency between image guidance and reality serves as a crucial factor in 3D scene reconstruction, many methods \cite{ni2025recondreamer, zhao2025recondreamer++, zhao2025drivedreamer4d, yan2025streetcrafter} treat video generation model as a reliable provider of novel-view guidance.
However, although video generation indeed provides general consistency in the exact input sequence (e.g. the car still goes straight in the same lane), the results are not only variant concerning high-frequency information such as details of building and vehicle exteriors, but also show significant changes between stitched video clips.
The detail variant nature of these methods degrades the quality of generated novel guidance and consequently limits scene reconstruction results.
Nevertheless, digital twin systems and autonomous driving simulation require a large asset library that needs to be obtained cost-effectively and efficiently, where the high GPU usage and slow generation time of video diffusion still largely prohibit StreetCrafter and other relevant works from facilitating real-world applications.

To get out of such a predicament, we believe the ultimate goal is to directly establish high consistency between generated images and real capture, instead of inside image sequences.
Inspired by CycleGAN-turbo and Pix2Pix-turbo \cite{parmar2024one} which achieve highly controlled image translation with consistent details, in this paper we present LidarPainter, a one-step diffusion model that effectively generates high-fidelity novel guidance from artifact rendering and LiDAR rendering for driving scene reconstruction.
By transferring the original video generation task to controlled image generation, LidarPainter achieves better quality and significantly lower resource consumption, enabling real-world implementations (e.g. can be running on a single RTX 2080 Ti GPU).


Given a set of driving scene capture, we first train the 3DGS \cite{kerbl20233d} representation for scene actors and background on early iterations with the original trajectory.
Once stabilized, we render images on new trajectories, which contain many artifacts and corrupted object structures because of the lack of supervision.
Meanwhile, we render LiDAR point cloud images on these novel views, which provide structural guidance.
The corrupted images and paired LiDAR images are then sent through LidarPainter's one-step diffusion model.
LidarPainter introduces a Latent Attention Fusion Module that can both take in LiDAR's structural information and preserve high frequency details, generating photorealistic images with removed artifacts and restored structure, which are made extra novel view supervision.
By continuing to train the scene gaussians with both original capture and novel view guidance, we obtain a driving scene reconstruction with improved quality and view range, which can be further extended with multiple rounds of novel view expansion.

Experiments show that LidarPainter can generate high-fidelity novel view guidance with structure consistency and detail accuracy (see Figure \ref{fig:intro}), achieving more reliable trajectory interpolation and extrapolation results compared with state-of-the-art methods for driving scene reconstruction.
In particular, our method is also superior in speed and resource efficiency, where LidarPainter is 7 $\times$ faster than the latest StreetCrafter with only one fifth of GPU memory required.
Moreover, LidarPainter further supports stylized generation using text prompts such as ``foggy'' and ``night'', allowing for a diverse expansion of the existing asset library.
Extensive ablation study shows the effectiveness of the LidarPainter network, which is promising for being adapted to other downstream tasks.

In summary, the principal contributions of this work include:
\begin{itemize}
\item We present LidarPainter, which to our best knowledge, is the first one-step diffusion model that can achieve better generation quality and consistency compared with video diffusion models trained on large driving datasets, enabling higher driving scene reconstruction quality. 
\item As far as we know, LidarPainter is one of the \textbf{fastest} and most \textbf{resource efficient} method for driving-scene novel-view generation, facilitating real-world applications.
\item We propose the Latent Attention Fusion Module in LidarPainter, which effectively achieves region-aware fusion generation with multiple inputs, empowering controlled generation for many other downstream tasks.
\end{itemize}

\section{Related Work}
\subsection{NeRF and 3D Gaussian Splatting (3DGS)}
Neural Radiance Fields (NeRF) and 3D Gaussian Splatting (3DGS) have revolutionized 3D scene modeling and rendering. After first introduction, NeRF \cite{mildenhall2021nerf} has advanced largely with further innovations \cite{kumar2025dynamode, zheng2025flow, wu2025pbr}. 3D Gaussian Splatting, detailed by \cite{kerbl20233d} and expanded upon in work \cite{sun2025splatflow, kwon2025efficient, lu2025bard} on dynamic 3D Gaussians, optimizes the rendering of Gaussian kernels for real-time applications. Other contributions \cite{chen2024mvsplat, Tian2025drive, tian2025uniforwardunified3dscene, Xie2026robotic, Hu2026multi} also underscore the ongoing enhancements and versatility of 3DGS in handling increasingly complex rendering tasks. Noting the advantages of 3D Gaussian Splatting in rendering speed and quality, our paper chooses it as a representation for modeling dynamic driving scenes.

\subsection{Driving Scene Reconstruction}
Both NeRF and 3DGS can be implemented for driving scene \cite{Tan_2025} reconstruction. While there are moving traffic participants such as pedestrians and vehicles, the scene cannot be modeled the same as in static scenes.

Existing techniques on dynamic scene reconstruction are mainly divided into two categories, one is to leverage time as an additional parameter to capture temporal variations in dynamic scenes \cite{attal2023hyperreel, fridovich2023k, song2023nerfplayer, huang2024textit}, but these methods do not show good support for scene manipulation.
The other way is to separately model scene background and moving traffic participants \cite{wu2023mars, yang2023emernerf, yang2023unisim, tonderski2024neurad, yan2024street, hess2025splatad, chen2025omnire}, so that the scene can be easily edited, but may result in high storage and rendering overhead in large scenes without optimization.

Specifically, StreetGaussians \cite{yan2024street} models the background and each moving object using separate Gaussian models. EmerNeRF \cite{yang2023emernerf} stratifies scenes into static and dynamic fields, each modeled with a hash grid. UniSim \cite{yang2023unisim} and NeuRAD \cite{tonderski2024neurad} utilize neural feature grids to model dynamic driving scenes with CNN renderer to enhance the ability of view extrapolation.
But without supervision on novel trajectories, the reconstruction quality of these methods is heavily restricted.

\subsection{Reconstruction With Diffusion Prior}
Advancements in image and video diffusion models \cite{rombach2022high, blattmann2023stable, yang2024cogvideox} have achieved highly controlled generation, making it possible to leverage diffusion prior as photorealistic guidance \cite{Gong2026diff, WANG2026111774, li2025one} for scene reconstruction.
DriveDreamer4D \cite{zhao2025drivedreamer4d} uses driving world model \cite{wang2024drivedreamer, zhao2025drivedreamer} to directly generate novel-view videos for training, improved reconstruction quality but still has corrupted novel renderings because of poor generation consistency.
StreetCrafter \cite{yan2025streetcrafter} trained a video diffusion model conditioned on LiDAR rendering to achieve precise camera control, but degrades on sparse LiDAR condition.
ReconDreamer \cite{ni2025recondreamer, zhao2025recondreamer++} trained a video restorer to recover clear views from novel view degraded images, but still suffers from high time and memory consumption, and limited quality as other video diffusion based methods.

Meanwhile, the potential of image diffusion prior is also explored.
DIFIX3D+ \cite{wu2025difix3d+} trained a diffusion model that takes in surround view guidance to restore novel views with extreme artifacts, but creates significant ghosting areas.
FreeSim \cite{fan2025freesim} treats this as an image enhancement task, but can only provide acceptable quality under progressive shifting with a step size of less than 1 meter, which is impossible for real-world applications because the size of scenes grows cubically.

\begin{figure*}[!t]\centering
  \includegraphics[width=1\linewidth]{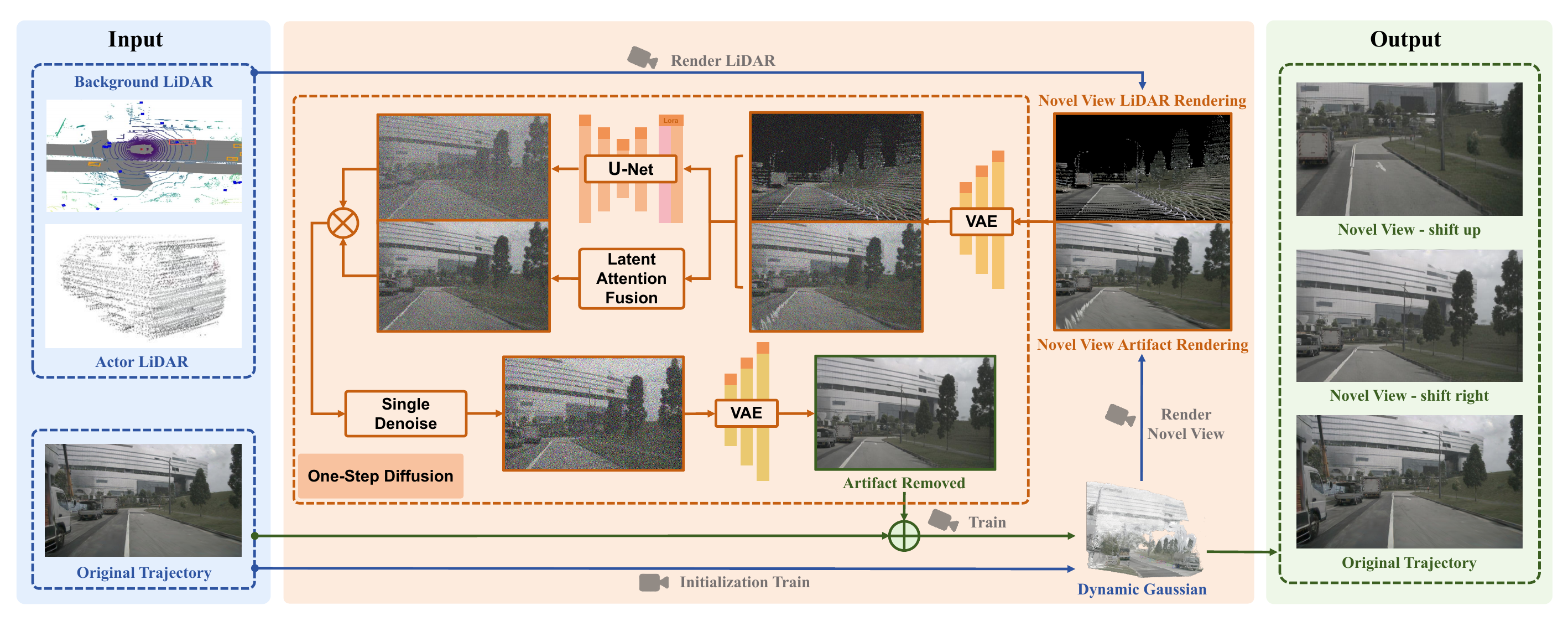}
  \caption{LidarPainter Reconstruction Pipeline.}
  \label{fig:pipeline}
\end{figure*}
\section{Method}
\subsection{Overview}
The general pipeline of driving scene reconstruction with LidarPainter is shown in Figure \ref{fig:pipeline}.
Given a set of driving scene capture, we first conduct an initialization train for dynamic 3DGS \cite{kerbl20233d} representation with the original trajectory.
Once stabilized, novel view images with artifacts will be rendered along with synced LiDAR point cloud images containing structural guidance, which are then sent through LidarPainter's one-step diffusion model.
LidarPainter generates photorealistic images with removed artifacts and restored structure, which are made extra novel view supervision.
By continuing to train the scene gaussians with both original capture and novel view guidance, we obtain a driving scene reconstruction with improved quality and view range, which can be further extended with multiple rounds of novel view expansion.

\subsection{Initialization}
First, we unify all data and prepare for LiDAR rendering on novel views.

\textbf{Unifying Data}
Driving datasets \cite{caesar2020nuscenes, sun2020scalability, xiao2021pandaset} are captured on multiple sensors including egocentric cameras, LiDAR and radars, which comes with different sampling frequency and coordinate system.
Therefore, it is necessary to unify all data for the same processing paradigm to ensure temporal and spatial consistency.
Specifically, LiDAR point cloud, ego poses, camera extrinsics and actor bounding box annotations are all transformed to world coordinate system.
Meanwhile, we use only LiDAR information that is synchronized with camera capture to avoid deviated actor positions.

\textbf{LiDAR Rendering}
LiDAR images are used both in training our diffusion model and scene reconstruction.
In a driving scene sequence with $N$ captured frames, the LiDAR points fall in $M$ actor bounding box annotations build actor point cloud $\{\mathbf{P}^{a}_{i}\}^{M-1}_{i=0}$, while the rest build background point cloud $\{\mathbf{P}^{b}_{i}\}^{N-1}_{i=0}$.
Given a camera pose $\mathbf{C^t}$ on frame time $\{t_i\}^{N-1}_{i=0}$, we aggregate the background LiDAR points $\{\mathbf{P}^{b}_{i}\}^{N-1}_{i=0}$ within a temporal window of size $l$ to form a unified point cloud $\mathbf{P}^{l}$ in the world coordinate system and perform point rasterization under the given camera pose $\mathbf{C^t}$ as in StreetCrafter \cite{yan2025streetcrafter} for background rendering. 
The coordinate of actor point cloud $\{\mathbf{P}^{a}_{i}\}^{M-1}_{i=0}$ built by complete capture sequence is transformed based on actor annotations on frame time $\{t_i\}^{N-1}_{i=0}$, and rendered separately.


\subsection{Novel View Guidance With One-Step Diffusion}
To improve scene reconstruction quality of driving scene captured on single trajectory, we leverage diffusion prior as guidance by building an efficient one-step diffusion model.
We build LidarPainter upon SD-Turbo \cite{sauer2024adversarial}, and the network structure is shown in Figure \ref{fig:pipeline}.
In this section we introduce the design of LidarPainter, while detailed ablations can be found in Ablation Studies.

\textbf{Preliminaries}
Diffusion Models (DM) \cite{sohl2015deep} are probabilistic models designed to learn a data distribution $p(x)$ by gradually denoising a normally distributed variable, which corresponds to learning the reverse process of a fixed Markov Chain of length $T$.
For image synthesis, the models can be interpreted as an equally weighted sequence of denoising autoencoders $\epsilon_\theta(x_t,t)$, which are trained to predict a denoised variant from noisy input $x_t$.
The simplified objective is:
\begin{equation}
    L_{D M}=\mathbb{E}_{x, \epsilon \sim \mathcal{N}(0,1), t}\left[\left\| \epsilon-\epsilon_{\theta}\left(x_{t}, t\right)\right\| _{2}^{2}\right] ,
\end{equation}
where $t$ is uniformly sampled from $\{1,...,T\}$.

In Latent Diffusion Models (LDM) \cite{rombach2022high}, the data \( x_t \) is encoded into a latent representation \( z_t \) via the perceptual compression encoder $\mathcal{E}$, while samples from $p(z)$ can be decoded to image space with a single pass through decoder $\mathcal{D}$.
LDM realizes the neural backbone $\epsilon_\theta(o,t)$ as a time-conditional UNet \cite{ronneberger2015u}, and the simplified objective is can be:
\begin{equation}
    L_{L D M}=\mathbb{E}_{\mathcal{E}(x), \epsilon \sim \mathcal{N}(0,1), t}\left[\left\| \epsilon-\epsilon_{\theta}\left(z_{t}, t\right)\right\| _{2}^{2}\right] .
\end{equation}

\textbf{Adding LiDAR Condition}
One-step diffusion has proven effective for image-to-image translation tasks \cite{sauer2024adversarial, parmar2024one, wu2025difix3d+}, but the task is much more demanding in generating highly consistent novel view guidance for 3D reconstruction than 2D style transfer.
DIFIX3D+ \cite{wu2025difix3d+} first trains a single-step diffusion model to remove novel-view artifacts for expanded guidance, but fails to restore fully corrupted targets such as lane lines.
Therefore, finding LiDAR renderings can provide strong structural guidance in novel views, we add LiDAR as image diffusion condition.

Given an artifact-corrupted novel-view rendering $I^A$ and corresponding LiDAR rendering $I^L$, we sent them separately through VAE encoder $\mathcal{E}$ to obtain latent representation $z$:
\begin{equation}
    z^A = \mathcal{E}(I^A); z^L = \mathcal{E}(I^L) ,
\end{equation}
which are then feed together to time-conditional UNet for predicting noise sample at timestep $t$:
\begin{equation}
    Z^{com}_t=\epsilon_{\theta}\left((z^A,z^L), t\right) .
\end{equation}

Normally, the noise sample is denoised conditioned on original noise sample.
But we find it insufficient with multiple inputs.
Therefore, we propose a Latent Attention Fusion Module to achieve region-aware fusion generation.

\textbf{Latent Attention Fusion}
While LiDAR rendering provides strong structural guidance in terms of fully corrupted targets, it is incapable of handling details such as small components and text.
In order to both preserve details from artifact images and restore corruption with LiDAR rendering, we conduct pixel-wise noise fusion.

To match the fused noisy residual $Z^{com}_t$, we fuse latent representation $z^A,z^L$ as:
\begin{equation}
    z^{com} = Att(z^A,z^L) \odot z^A + (1-Att(z^A,z^L)) \odot z^L ,
\end{equation}
where $\odot$ denotes element-wise multiplication and $Att$ indicates a trainable 2D convolutional module that generates pixel-wise fusion weights.

Then the predicted noise sample $Z^{com}_t$ is denoised conditioned on fused latent $z^{com}$ by:
\begin{equation}
    z^D = \frac{\sqrt{\alpha_{t-1}} \cdot \beta_t}{\beta_{t-1}} \cdot Z^{com}_t + \frac{\sqrt{\alpha_t} \cdot \beta_{t-1}}{\beta_t} \cdot z^{com} + \sigma_t \cdot \epsilon_t ,
\end{equation}
where $\alpha_t$ and $\beta_t$ are the strength of original signal and added noise, $\epsilon_t$ is perturbation noise, $\sigma_t$ is the standard deviation of the Gaussian noise at timestep $t$ (controlled by the schedule).

Finally, the denoised latent $z^D$ is decoded by $\mathcal{D}$ to image space:
\begin{equation}
    I^O = \mathcal{D}(z^D).
\end{equation}

\textbf{Losses}
We use LoRA adapters \cite{hu2022lora} in VAE and U-Net modules, and train the Latent Attention Fusion Module. We also add skip connections between the Encoder and Decoder networks as in pix2pix-turbo \cite{parmar2024one} for fine-grained details.

For reconstruction loss, we use L2 difference and Structural Similarity Index Measure (SSIM).
For perceptual loss, we use LPIPS \cite{zhang2018unreasonable} and GAN loss \cite{parmar2024one}.

The final loss term is: 
\begin{equation}
    \mathcal{L} = 0.2\mathcal{L}_{LPIPS} + 0.6\mathcal{L}_2 + 0.4\mathcal{L}_{SSIM} + \mathcal{L}_{GAN} .
\end{equation}

\subsection{Dynamic Scene Reconstruction}
In order to have strong scene editing ability, we separately model scene background and moving traffic participants for dynamic scene reconstruction.

\textbf{Basic Models}
In LidarPainter, each Gaussian primitive $\mathcal{G}$ is parameterized by properties $\{\mu, \Sigma, \alpha, c\}$ following the original 3DGS \cite{kerbl20233d}, where the covariance matrix $\Sigma$ can be decomposed into a scaling factor $s \in \mathbb{R}^3_+$ and a rotation quaternion $r \in \mathbb{R}^4$.

While background model is already in the world coordinate system, moving objects are separately optimized as local models.
Given the SE(3) pose $T_v = (R_v, t_v)$ of an vehicle $v$, where $R_v$ is a 3x3 rotation matrix from SO(3) and $t_v$ is a 3x1 translation vector in $\mathbb{R}^3$, the Gaussian primitives $\mathcal{G}_v$ can be transformed to world coordinate system by:
\begin{equation}
    \hat{\mu}_v = R_v\mu_v + t_v, \space \hat{R}_v = R_vr^{mat}_v ,
\end{equation}
where $\hat{\mu}_v$ and $\hat{R}_v$ indicate the position and rotation in world coordinate system, and $r^{mat}_v$ denotes rotation matrix of $\mathcal{G}_v$.

Given a camera pose, we compute the colors by blending a set of ordered Gaussians $\mathcal{N}$ overlapping the pixel as in 3DGS \cite{kerbl20233d}: 
\begin{equation}
\mathbf{C}=\sum_{i\in\mathcal{N}}\mathbf{c}_{i}a_i\prod\limits_{j=1}^{i-1}(1-a_j),
\end{equation}
where $a_i$ is given by evaluating a 2D Gaussian with covariance $\sum$ multiplied with a learned per-Gaussian opacity.

\textbf{Reconstruction Process}
We first train the background and object Gaussians on the original trajectory to stabilize their 3D representations until sample iteration $T_s$.
Then we render novel view images and use LidarPainter's one-step diffusion to generate novel guidance.
We continue training the Gaussians to final iteration $T_e$ using both input images and novel guidance, with a sampling probability of $p_{novel}$ for novel guidance.
The process of novel view expansion can be repeated for progressively improving novel-view quality or keeping increasing view range.

\textbf{Losses}
For reconstruction loss, we use L1 difference and Structural Similarity Index Measure (SSIM).
For perceptual loss, we use LPIPS.
Extra loss $\mathcal{L}_g$ for sky mask and moving objects regularization \cite{yan2025streetcrafter} is also applied.
The loss term for original trajectory is:
\begin{equation}
    \mathcal{L}_{original} = \mathcal{L}_1 + 0.2\mathcal{L}_{SSIM} + \mathcal{L}_{LPIPS} + \mathcal{L}_g ,
\end{equation}
and novel view loss is:
\begin{equation}
    \mathcal{L}_{novel} = \lambda_{novel}(\mathcal{L}_1 + 0.1\mathcal{L}_{SSIM} + \mathcal{L}_{LPIPS} + \mathcal{L}_g) .
\end{equation}

\section{Experiments}
\subsection{Experimental Setup}
In this section, we conduct extensive experiments to prove the superiority and rationality of LidarPainter design.

\textbf{Datasets and Metrics}
For quantitative experiments, we train and evaluate the models on Waymo Open Dataset \cite{sun2020scalability}, PandaSet \cite{xiao2021pandaset} and nuScenes dataset \cite{caesar2020nuscenes}.
Our one-step diffusion model is trained on 100 sequences of Waymo Open Dataset and nuScenes dataset including 5000 frames.
We use 15 sequences from Waymo, 5 sequences from PandaSet (as in StreetCrafter \cite{yan2025streetcrafter}) and 10 sequences from nuScenes to test novel view synthesis ability.

To evaluate the quality of scene reconstruction, we use the peak signal-to-noise ratio (PSNR) and perceptual distance (LPIPS) \cite{zhang2018unreasonable}.
To evaluate the quality of novel view generation, we use fréchet inception distance (FID) \cite{heusel2017gans}.

\textbf{Implementation Details}
LidarPainter's diffusion model is trained for 36 hours using Adam \cite{kingma2014adam} optimizer, with each image resized to $512 \times 512$.
During inference, the generated image size is $1024 \times 576$.
For original trajectory, we uniformly sample half of the images in each sequence as the testing frames and use the remaining for training the same as in StreetCrafter.
We test LidarPainter with single round of novel sample.
Sample iteration $T_s$ and final iteration $T_e$ for training are set to 7000 and 30,000.
Sampling probability $p_{novel}$ for novel guidance is 0.4, and $\lambda_{novel}$ is 0.2.
We use the same $\lambda_{novel}$ and sampling probability $p_{novel}$ for LidarPainter, StreetCrafter and DIFIX during experiments, and others all follow their original settings.
We use the same artifact and LiDAR rendering input for LidarPainter and StreetCrafter, while providing DIFIX with ground-truth training views as reference.
All results (including the training for diffusion model) are reported running on a single NVIDIA A100 80GB GPU.

\begin{figure*}[!t]\centering
  \includegraphics[width=1\linewidth]{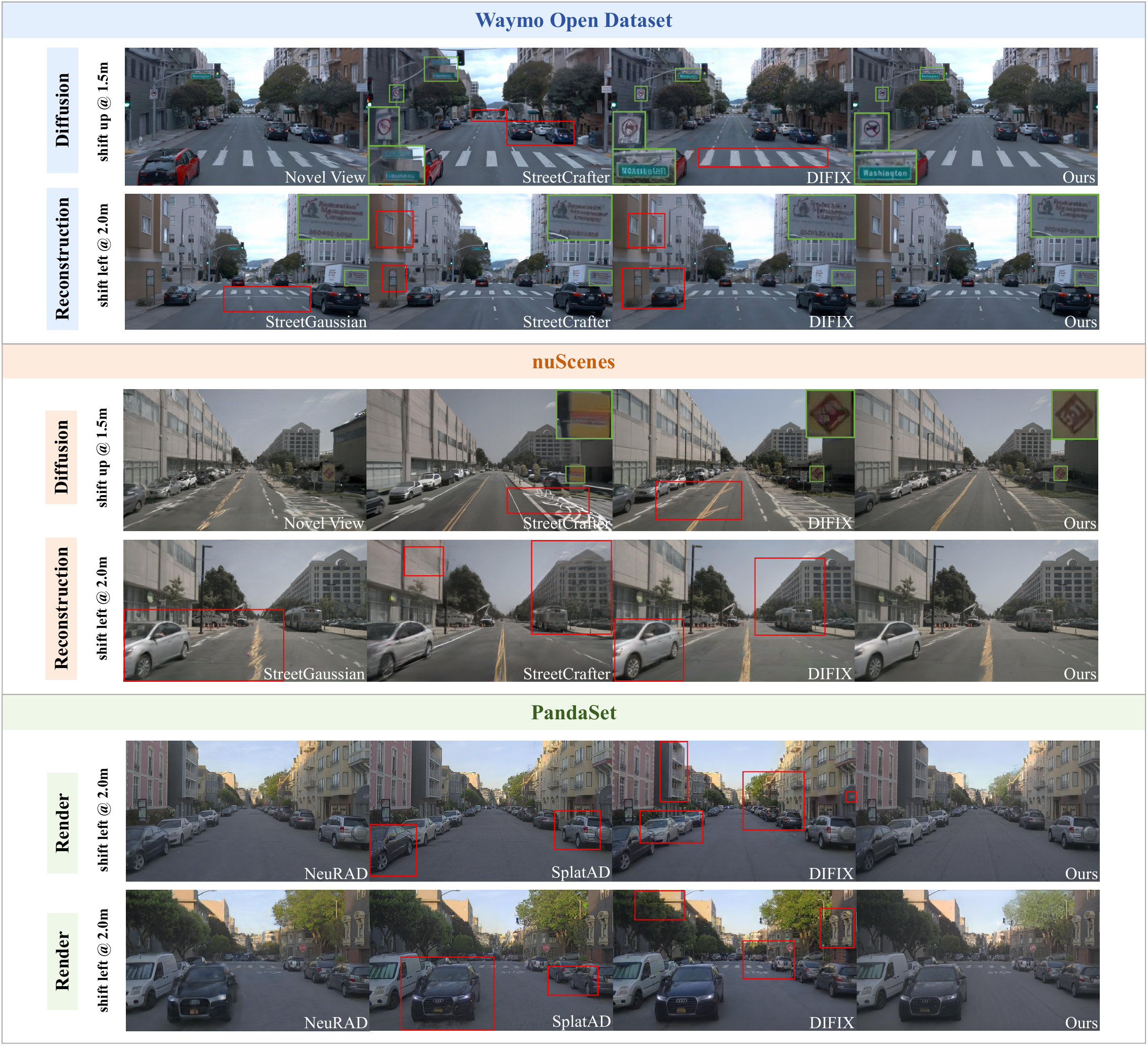}
  \caption{Qualitative comparisons of image generation and 3D scene reconstruction results on different scenes.}
  \label{fig:qualitative}
\end{figure*}
\subsection{Comparison With Other State-of-the-art Methods}
We compare LidarPainter with other state-of-the-art methods by evaluating the rendered quality of trajectory interpolation and extrapolation in reconstructed scenes.
We report PSNR and LPIPS on trajectory interpolation, and evaluate FID on trajectory extrapolation since there is no ground truth for novel views.
We compare with EmerNeRF \cite{yang2023emernerf}, UniSim \cite{yang2023unisim} and NeuRAD \cite{tonderski2024neurad} (NeRF-based methods), StreetGaussians \cite{yan2024street}, StreetCrafter \cite{yan2025streetcrafter} and SplatAD \cite{hess2025splatad} (3DGS-based methods).
We also compare with the DIFIX3D+ \cite{wu2025difix3d+}, a static scene reconstruction method equipped with the first-tier artifact removal module, by replacing LidarPainter with DIFIX in our reconstruction pipeline.


\begin{table}
\centering
\resizebox{\columnwidth}{!}{
\begin{tabular}{lccccc}
\toprule
\multirow{2}{*}{Methods} & \multicolumn{2}{c}{Interpolation} & \multicolumn{3}{c}{Lane Shift @ FID$\downarrow$} \\ 
\cmidrule(lr){2-3} \cmidrule(lr){4-6} 
 & PSNR$\uparrow$ & LPIPS$\downarrow$ & side 2m & side 3m & up 1.5m \\
 \midrule
EmerNeRF (2023) & 26.09 & 0.199 & 90.17 & 109.14 & 87.73 \\
StreetGaussians (2024) & \underline{30.82} & 0.146 & 73.96 & 94.68 & 71.43 \\
StreetCrafter (2025) & 30.01 & 0.072 & 57.12 & 69.35 & 58.04 \\ 
DIFIX (2025) & 29.13 & 0.082 & 58.17 & 77.80 & 63.19 \\
LidarPainter (Ours) & \textbf{31.04} & \textbf{0.067} & \textbf{53.71} & \textbf{67.52} & \textbf{52.53} \\
\bottomrule
\end{tabular}
}
\caption{Quantitative results on Waymo Open Dataset.}
\label{tab:waymo}
\end{table}


\begin{table}
\centering
\resizebox{\columnwidth}{!}{
\begin{tabular}{lccccc}
\toprule
\multirow{2}{*}{Methods} & \multicolumn{2}{c}{Interpolation} & \multicolumn{3}{c}{Lane Shift @ FID$\downarrow$} \\ 
\cmidrule(lr){2-3} \cmidrule(lr){4-6} 
 & PSNR$\uparrow$ & LPIPS$\downarrow$ & side 2m & side 3m & up 1.5m \\
 \midrule
NeuRAD (2024) & 27.24 & 0.109 & 73.62 & 95.40 & 70.31 \\
StreetGaussians (2024) & 27.56 & 0.157 & 77.84 & 96.32 & 73.73 \\
SplatAD (2025) & \textbf{28.02} & 0.099 & 69.25 & 87.62 & 68.83 \\
StreetCrafter (2025) & 27.16 & 0.097 & 64.34 & 79.01 & 61.03 \\ 
DIFIX (2025) & 26.98 & 0.101 & 65.82 & 86.31 & 66.69 \\
LidarPainter (Ours) & \underline{27.83} & \textbf{0.072} & \textbf{60.32} & \textbf{69.41} & \textbf{57.33} \\
\bottomrule
\end{tabular}
}
\caption{Quantitative results on nuScenes Dataset.}
\label{tab:nuscenes}
\end{table}


\begin{table}
\centering
\resizebox{\columnwidth}{!}{
\begin{tabular}{lccccc}
\toprule
\multirow{2}{*}{Methods} & \multicolumn{2}{c}{Interpolation} & \multicolumn{3}{c}{Lane Shift @ FID$\downarrow$} \\ 
\cmidrule(lr){2-3} \cmidrule(lr){4-6} 
 & PSNR$\uparrow$ & LPIPS$\downarrow$ & side 2m & side 3m & up 1.5m \\
 \midrule
UniSim (2023) & 25.61 & 0.121 & 74.34 & 94.17 & 74.11 \\
NeuRAD (2024) & 27.19 & 0.106 & 63.45 & 87.42 & 61.51 \\
StreetGaussians (2024) & 27.40 & 0.117 & 69.84 & 89.31 & 66.73 \\
SplatAD (2025) & \textbf{27.51} & 0.103 & 67.82 & 87.42 & 65.01 \\ 
StreetCrafter (2025) & 27.39 & 0.098 & 66.74 & 80.21 & 63.96 \\ 
LidarPainter (Ours) & \underline{27.50} & \textbf{0.086} & \textbf{61.16} & \textbf{77.35} & \textbf{60.75} \\
\bottomrule
\end{tabular}
}
\caption{Quantitative results on PandaSet.}
\label{tab:pandaset}
\end{table}

\textbf{Qualitative Comparison}
Qualitative visual results of image generation and 3D scene reconstruction are shown in Figure \ref{fig:qualitative}, with framed parts for comparison.

For the task of generating novel view guidance with artifact rendering, the latest diffusion methods StreetCrafter and DIFIX both destroy text regions and signs.
StreetCrafter also generates corrupted vehicles and even cars that are inexistent in the input rendering, while DIFIX fails to restore deformed lane lines (even with ground truth view as reference).
Only our method can simultaneously fix lane lines, remove artifacts and preserve details.

For the task of 3D scene reconstruction, StreetGaussian solely trained on original trajectory creates highly deformed lane lines in novel trajectory due to the lack of supervision.
With inconsistent novel guidance, StreetCrafter and DIFIX still fail in text regions, resulting in new artifacts on vehicles and buildings, while our method provides the smoothest reconstructed results on novel trajectory.

We also test LidarPainter by implementing it for post render processing at rendering stage of reconstructed scenes as in DIFIX3D+ \cite{wu2025difix3d+}.
In Figure \ref{fig:qualitative}, results on PandaSet shows that both DIFIX and LidarPainter can reduce artifacts comparing with single NeRF and 3DGS rendering.
However, DIFIX still generates corrupted structure and transparent artifact, which contradicts the original intention of using post-processing to improve image quality.

\textbf{Quantitative Comparison}
We report quantitative results of trajectory interpolation (original trajectory) and extrapolation (lane shift, novel trajectory) in Table \ref{tab:waymo}, \ref{tab:nuscenes} and \ref{tab:pandaset}.

In trajectory interpolation, although methods such as StreetGaussians and SplatAD are solely trained on original trajectory using ground truth training views (with around 10000 more iterations), our method can still have competitive results with even better PSNR on Waymo Open Dataset.
This means that correct novel guidance will improve the reconstruction quality on the original trajectory. 

In trajectory extrapolation, we test the novel view quality on side (left and right) and up shifted cameras.
Novel guidance is important in reducing artifacts and restore corruptions, thus methods trained only on original trajectory degrades significantly.
Meanwhile, LidarPainter generates more consistent novel supervision than StreetCrafter and DIFIX, resulting in the lowest FID.

\textbf{Speed And Resource Efficiency}
To obtain abundant 3D assets for autonomous driving simulation systems, the speed and resource efficiency for a reconstruction pipeline are equally important as its quality.
While reconstruction module can be easily changed, we compare the speed and consumption of LidarPainter with StreetCrafter's video diffusion and DIFIX3D+'s image diffusion.
The result is shown in Table \ref{tab:resource}, where LidarPainter shows clear advantage.

In conclusion, LidarPainter is superior in quality, speed, and resource efficiency as mentioned above, showing more possibility in real-world applications.

\begin{table}
\centering
\begin{tabular}{ccccc}
  \toprule
  Method    & StreetCrafter & DIFIX${}^*$ & OURS\\
  \midrule
  Inference (s/frame)     & 6.27 & 1.06 & \textbf{0.87}\\
  GPU Usage (GB)      & 55.3 & 10.9 & \textbf{10.5}\\
  \bottomrule
\end{tabular}
\caption{Comparison of performance and time consumption computed on rendering resolution 1024 $\times$ 576 (${}^*$only single shifting time in progressive update).}
\label{tab:resource}
\end{table}

\subsection{Ablation Studies}
We conduct ablation to validate the effectiveness of LidarPainter design.

\begin{table}
\centering
\begin{tabular}{cccccc}
  \toprule
  Input & \multicolumn{5}{c}{Ablations}\\
  \midrule
  Artifact&            &            & \checkmark & \checkmark & \checkmark\\
  LiDAR   & \checkmark & \checkmark &            &            & \checkmark\\
  GT      & \checkmark &            & \checkmark &            & \\
  \midrule
  FID$\downarrow$ & 109.72 & 96.94 & 79.63 & 73.31 & \textbf{66.84} \\
  \bottomrule
\end{tabular}
\caption{Ablation on input guidance. Artifact, LiDAR and GT denotes artifact rendering, LiDAR rendering and ground truth training view.}
\label{tab:ablation_input}
\end{table}
\textbf{Input Guidance}
We analyze the influence of different input guidance for LidarPainter on view generation quality.
The comparison of novel view FID is reported in Table \ref{tab:ablation_input}.
Only using artifact rendering for training will result in a de-blur model that can successfully restored blurred areas but cannot restore corrupted structure, while single LiDAR input maintains structure yet provides poor image quality.
Although the intention is to use the best guidance, adding nearby ground truth image as reference (similar to DIFIX3D+ \cite{wu2025difix3d+}) will lead to even worse results with extra artifacts, because the model cannot mix the different views properly without spatial guidance.
Use paired (same view) artifact rendering and LiDAR rendering can easily use both guidance and generate better images.

\begin{figure}
   \centering
   \includegraphics[width = 0.9\columnwidth]{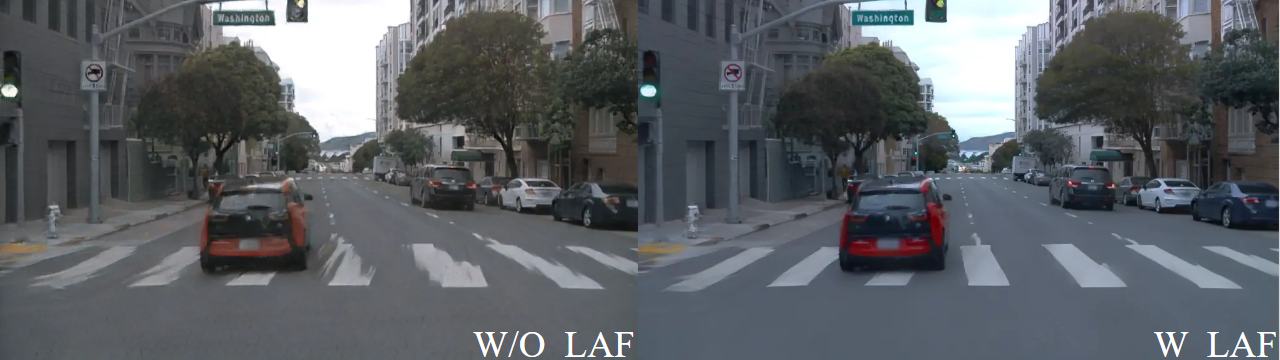}
   \caption{Ablation on Laten Attention Fusion (LAF).}
   \label{fig:ablation}
\end{figure}
\textbf{Latent Attention Fusion}
We analyze the necessity of our latent attention fusion module.
As shown in Figure \ref{fig:ablation}, without Laten Attention Fusion module, the diffusion process is more of a de-blur process that removes surrounding artifacts, but fails in restoring torn lane lines.
The Laten Attention Fusion module can correctly use LiDAR guidance to restore highly corrupted structure and preserve its original de-blur ability for other regions with high-frequency information such as text and vehicles.

\begin{figure}
   \centering
   \includegraphics[width = 0.9\columnwidth]{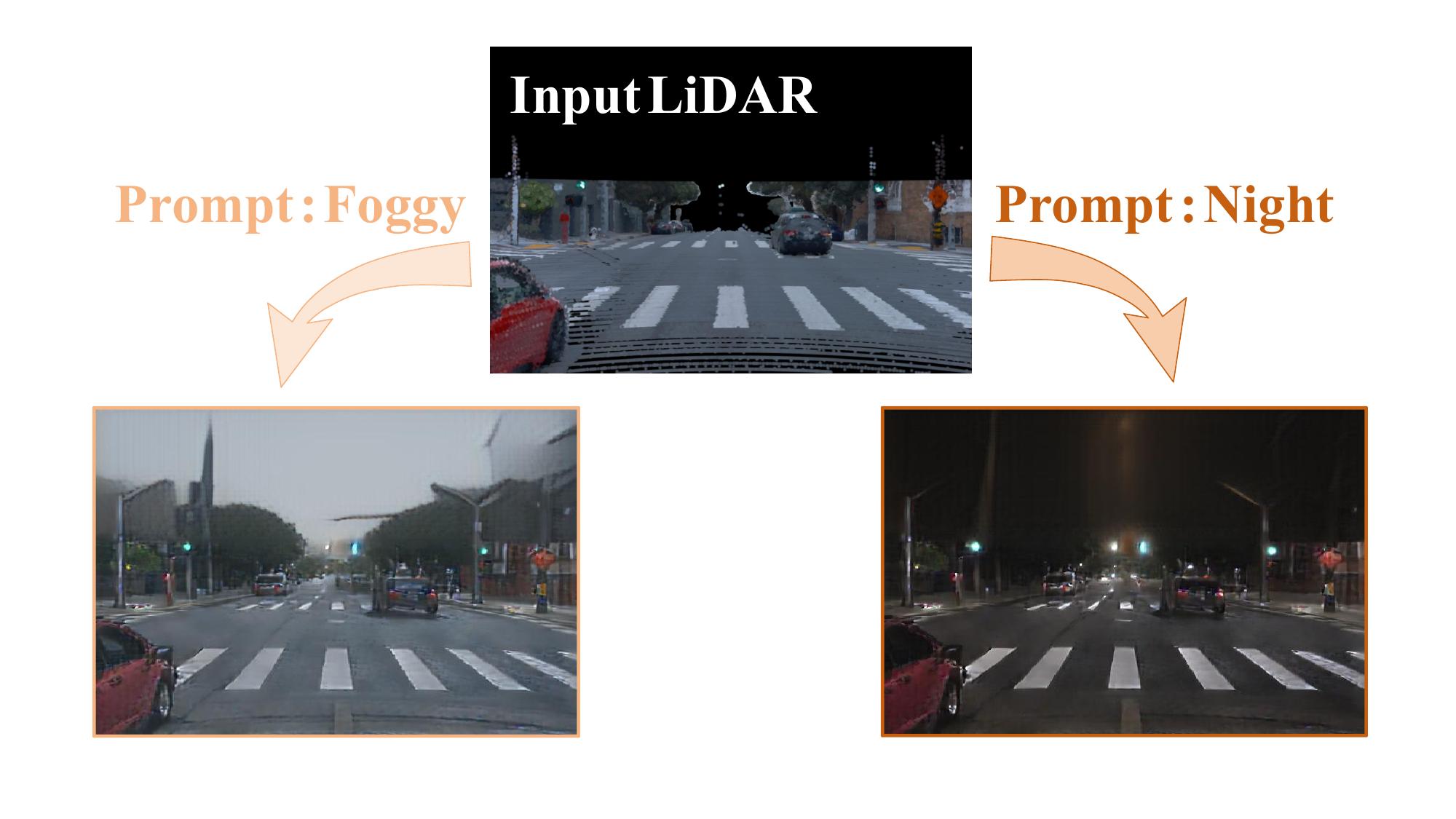}
   \caption{Prompted generation of LidarPainter.}
   \label{fig:style}
\end{figure}
\subsection{Stylized Generation}
LidarPainter also supports stylized generation based on text prompt by conditioning the latent representation $z$ on CLIP \cite{radford2021learning} embeddings.
We trained another LidarPainter model using 600 paired images of original capture, foggy and night scenes (generated by CycleGAN-turbo \cite{parmar2024one}) with corresponding text prompts.
As shown in Figure \ref{fig:style}, with the same input LiDAR rendering, LidarPainter can generate views on different circumstances according to text prompts and maintain the original structure given by LiDAR.
This allows for a diverse expansion of the existing asset library without additional real-world capture, which could be expensive and time-consuming.

\section{Conclusion}
In this paper, we present LidarPainter, a one-step diffusion model that recovers consistent driving views from sparse LiDAR condition and artifact-corrupted renderings, enabling high-fidelity lane shifts in dynamic street scene reconstruction.
We present Latent Attention Fusion module implemented in one-step diffusion model to effectively use both structural information in LiDAR and details in artifact rendering, providing a promising solution for controlled image generation with multiple inputs.
LidarPainter is superior in speed, quality, and resource efficiency, which also supports stylized generation using text prompts, allowing for a diverse expansion of the existing asset library.

\section{Acknowledgments}
This work is done during an internship at 51WORLD, and is supported by the National Natural Science Foundation of China No.62302167, 62472282, 72192821, U23A20343; Shanghai Sailing Program 23YF1410500; Young Elite Scientists Sponsorship Program by CAST YESS20240780; the Chenguang Program of Shanghai Education Development Foundation and Shanghai Municipal Education Commission 23CGA34; Natural Science Foundation of Chongqing CSTB2023NSCQ-MSX0137; the Fundamental Research Funds for the Central Universities (project number: YG2023QNA35).

\bibliography{aaai2026}

\end{document}